\documentclass[fleqn,10pt]{wlscirep}

\usepackage[utf8]{inputenc}
\usepackage[T1]{fontenc}
\usepackage{lipsum}
\usepackage[version=4]{mhchem}
\usepackage{siunitx}
\DeclareSIUnit\Molar{M}
\usepackage{array}
\newcolumntype{P}[1]{>{\centering\arraybackslash}p{#1}}
\usepackage[colorinlistoftodos]{todonotes}
\usepackage{placeins}
\usepackage{graphicx}
\usepackage{placeins}
\usepackage{tabularx}
\usepackage{amsmath}
\graphicspath{{Figures/}}
\usepackage{flushend}
\usepackage{multirow}
\usepackage{booktabs}
\usepackage{soul}
\usepackage{tikz}
\usepackage{subcaption}
\usetikzlibrary{shapes.geometric, arrows}
\pgfarrowsdeclarecombine{blah}{blah}{diamond}{diamond}{stealth}{stealth}

\title{Proprioceptive Sonomyographic Control: A novel method of intuitive proportional control of multiple degrees of freedom for upper-extremity amputees}

\author[1]{Ananya S. Dhawan}
\author[2,3,+]{Biswarup Mukherjee}
\author[2,+]{Shriniwas Patwardhan}
\author[2]{Nima Akhlaghi}
\author[4]{Gyorgy Levay}
\author[5]{Rahsaan Holley}
\author[2,3]{Wilsaan Joiner}
\author[2,3,5]{Michelle Harris-Love}
\author[2,3,*]{Siddhartha Sikdar}
\affil[1]{Department of Computer Science, George Mason University, Fairfax, VA, 22030, USA}
\affil[2]{Department of Bioengineering, George Mason University, Fairfax VA, 22030, USA}
\affil[3]{Center for Adaptive Systems of Brain-Body Interactions, Fairfax, VA 22030}
\affil[4]{Infinite Biomedical Technologies, Baltimore, MD 21202, USA}
\affil[5]{MedStar National Rehabilitation Hospital, Washington, DC 20010, USA}

\affil[*]{ssikdar@gmu.edu}
\affil[+]{these authors contributed equally to this work}

\begin{abstract}

Technological advances in multi-articulated prosthetic hands have outpaced the methods available to amputees to intuitively control these devices. Amputees often cite difficulty of use as a key contributing factor for abandoning their prosthesis, creating a pressing need for improved control technology. A major challenge of traditional myoelectric control strategies using surface electromyography electrodes has been the difficulty in achieving intuitive and robust proportional control of multiple degrees of freedom. In this paper, we describe a new control method, proprioceptive sonomyographic control that overcomes several limitations of myoelectric control. In sonomyography, muscle mechanical deformation is sensed using ultrasound, as compared to electrical activation, and therefore the resulting control signals can directly control the position of the end effector. Compared to myoelectric control which controls the velocity of the end-effector device, sonomyographic control is more congruent with residual proprioception in the residual limb. We tested our approach with 5 upper-extremity amputees and able-bodied subjects using a virtual target achievement and holding task. Amputees and able-bodied participants demonstrated the ability to achieve positional control for 5 degrees of freedom with an hour of training. Our results demonstrate the potential of proprioceptive sonomyographic control for intuitive dexterous control of multiarticulated prostheses.

\end{abstract}
\begin{document}
\flushbottom
\maketitle
\thispagestyle{empty}

\section*{Introduction}

Currently, there are approximately 600,000 individuals living with upper limb loss in the US. Upper extremity amputations most commonly occur in working age adults as a result of trauma~\cite{ziegler2008estimating,Dillingham1998}, and frequently affect the dominant extremity, leading to significant impacts on activities of daily living. The most common upper extremity amputation involving a wrist disarticulation or higher occurs at the transradial level (57\%)~\cite{Esquenazi1996}. Despite the enormous investment of resources in the development of new multi-articulated upper limb prostheses, a large proportion of upper extremity amputees discontinue use of their prosthesis~\cite{Biddiss2007Abandon,Ostlie2012,McFarland2010}. Prosthetic non-wear or part-time use has been reported in 20\% of the adult upper limb amputee population and rejection rates for upper limb prosthetic users are staggering, with reported figures ranging from 35-45\% for myoelectric and cable controlled systems~\cite{Biddiss2007Abandon}. Dissatisfaction with prosthesis technology is strongly associated with rejection, and 88\% of non-users reported the systems as being “too difficult or tiring” to use~\cite{Biddiss2007}.  However, 74\% of those who have abandoned their upper limb prostheses, stated that they would reconsider prosthetic use if technological advancements were made to improve their functionality and usability~\cite{Biddiss2007}.  Therefore, there is a significant need for better technological solutions to improve the function and quality of life of upper limb amputees.

While significant advances have been made in the electromechanical design of multiarticulated dexterous prostheses, methods enabling  amputees to intuitively control these devices is still lacking. Advanced myoelectric prosthetic hands either utilize two surface electromyography (sEMG) electrodes to record electrical activity from flexor and extensor muscles of the residuum, or use sophisticated machine learning-based systems, such as pattern recognition and regression~\cite{amsuess_multi-class_2015,stango2015spatial,hahne2014linear}, using multiple electrodes~\cite{chu2006real,Amsuess2015,Hargrove2010}. sEMG signals have poor signal-to-noise characteristics and hence limited amplitude resolution, making it challenging for users to accurately achieve graded levels of sEMG amplitude. Furthermore, sEMG signals suffer from random fluctuations of high amplitude, especially with dry electrodes~\cite{Clancy2002,Daley2012,Fillauer1989}. Finally, sEMG signals have limited specificity for deep contiguous multi-compartmental muscles because of cross talk~\cite{Kong2010,VanDuinen2010,VanDuinen2009,McIsaac2007}, and differentiation between individual digit and joint motions is challenging based on electrical activity recorded through the overlying tissues. Due to these limitations, in the most commonly used myoelectric control strategy, called direct control, sEMG amplitude is used to proportionally drive joint velocity instead of joint position. In velocity-based control, muscle contraction amplitude is mapped to a single movement, where flexion and extension dictate the movement direction while muscle inactivity halts joint movement. While this scheme allows for smooth joint movements with noisy sEMG amplitude measurements, the use of sEMG activation to create an artificial velocity control signal is not congruent with proprioceptive feedback from residual muscles. Proprioception is dependent on the mechanical movement of muscles, tendons and associated fascia, rather than the level of electrical activation~\cite{Doubler1984a,Doubler1984b}. Thus, velocity-based control severely limits a patient’s ability to achieve dexterous manipulation. Amputees often grip objects with excessive force and velocity as compared to able-bodied individuals~\cite{VanDijk2016}. As a result, many amputee users prefer body powered hooks over myoelectric systems, as the cable tension provides congruent sensory feedback for positional control and thus is more intuitive to use~\cite{Biddiss2007Abandon,Carey2015}.

Another significant challenge is to intuitively select among different grips in the terminal device. Advanced multiarticulated hands, such as the bebionic (Ottobock, GmbH) and i-LIMB (Touch Bionics, LLC)~\cite{Waryck2011} offer a number of different grip patterns (e.g., power, key, precision pinch, index point, three-jaw chuck grips). Current advanced hands use a method called mode switching that requires special sEMG triggers, e.g., co-contraction of both flexors and extensors in order to switch between different grasps. In the absence of muscle pairs in amputees, these functions are mapped to a set of substituted musculature or in some cases physical buttons on the prosthetic hand resulting in an unintuitive control paradigm. In addition, subjects experience periods of confusion with respect to the current operating mode of the limb, as well as the next mode in the control sequence. Not surprisingly, amputee subjects report that they strongly dislike mode switching~\cite{Hargrove2010}. A major limitation of current systems is the level of conscious attention required by the amputee user, and the lack of intuitive control~\cite{Carey2015}. Thus, there is an urgent need for a more intuitive command mechanism to control a lifelike prosthesis \cite{Kyberd2011,Atkins1996}. 

New EMG decoding methods are being actively researched to improve functionality~\cite{Jiang2014,Amsuess2015}. Most notably, pattern recognition algorithms are being developed to decode user intention from recorded temporal sequences of multi-channel sEMG signals.~\cite{Daley2012,Tenore2009,Li2010,Englehart1999,Englehart2003,Farrell2007,Smith2011,Scheme2011}. While pattern recognition is able to classify the intended grasp end-state with high accuracy, the ability of amputees to translate classification accuracy to intuitive real-time control with multiple degrees of freedom is limited~\cite{Simon2011}. In real life usage, pattern recognition lacks robustness and requires frequent retraining. To address these limitations of sEMG-based myoelectric control, other invasive strategies, such as implantable myoelectric systems~\cite{Pasquina2015,Weir2009,kuiken2009targeted}, targeted muscle reinnervation~\cite{kuiken2009targeted} and peripheral implant~\cite{Micera2009,Navarro2005} strategies, are being actively explored. While these methods can overcome many of the limitations described above, implanted devices and surgical procedures are associated with risks and side effects and may not be available to most amputees. Thus, there continues to be a need for a robust noninvasive strategy that can provide intuitive real-time proprioceptive control over multiple degrees of freedom, enabling amputees to make full use of advanced commercial hands.

In recent years, sonomyography or ultrasound-based sensing of mechancical muscle contractions, is being actively pursued as an noninvasive alternative to myoelectric control~\cite{akhlaghi2016real}. Sonomyography overcomes many of the limitations of current myoelectric control. Ultrasound imaging can spatially resolve individual muscles, including those deep inside the tissue, and detect dynamic activity within different functional compartments in real-time while being non-invasive. Sonomyography has been shown to be useful for detecting individual finger positions~\cite{castellini2012using,sikdar2014}, along with other complex muscle deformation patterns~\cite{hodges2003measurement}. Previous research has also shown that ultrasound techniques could be used for real-time classification of hand movements by predicting forearm muscle deformation patterns in able-bodied individuals~\cite{akhlaghi2016real,sikdar2014} as well as a trans-radial amputee~\cite{baker2016real} lending support to future prosthetic control applications. In this paper, we propose a new sonomyographic control strategy to classify volitional motion intent and further extend our paradigm to enable intuitive, position-based proportional control over multiple degrees-of-freedom. Since our strategy relies on mechanical muscle contractions and relaxations that are congruent with underlying proprioceptive feedback in the residual limb, we refer to our strategy as \textit{proprioceptive sonogmyographic control}. We validate our techniques on able-bodied subjects and apply it to five upper-extremity amputees one of whom is a congenital amputee. We asked the participants to perform predefined hand motions while ultrasound images of their forearm muscle movements were captured using a portable, commercial ultrasound system. We extracted representative ultrasound images from the collected ultrasound data and perform leave-one-out validation to quantify prediction accuracy. Participants were then asked to perform the same hand motions in real-time while being shown an on-screen cursor that moved up or down in proportion to their muscle movements. A series of targets were presented and the participant's ability to perform graded control using forearm muscle deformation was measured. The goals of this work are 1) to determine the ability of upper-extremity amputees to perform different hand motions using our \textit{proprioceptive sonomyographic control} strategy, with minimal training and 2) to determine the accuracy and stability with which amputees can perform these motions in a proportional manner.


\section*{Methods}
\label{sec:methods}

\subsection*{Subjects and experimental setup}
\label{sec:methodsSubjects}

We recruited four unilateral and one bilateral, upper-extremity amputees at the MedStar National Rehabilitation Hospital (NRH) and George Mason University (GMU). All of the amputee participants were at the time using electrically powered, myoelectric prostheses with varying levels of proficiency. Subject-specific amputation details and demographics are available in Table \ref{table:amp-demographics}. Additionally, five able-bodied subjects were recruited and served as a control group for this study. Demographics for able-bodied participants are listed in Table \ref{table:able-demographics}. All procedures described in this work were approved by the respective Institutional Review Boards at NRH and GMU. All subjects provided written, informed consent prior to participating in the study, and were also compensated for their participation.

\begin{table}[!h]
\centering
\caption{Demographics and amputation details of amputee subjects}
\label{table:amp-demographics}
\begin{tabular*}{\textwidth}{P{0.04\textwidth}P{0.04\textwidth}P{0.04\textwidth}P{0.15\textwidth}P{0.15\textwidth}P{0.45\textwidth}}
\toprule
\textbf{\begin{tabular}[c]{@{}c@{}}Subject\\ ID\end{tabular}} & \textbf{Sex} & \textbf{Age} & \textbf{\begin{tabular}[c]{@{}c@{}}Years since \\ amputation\end{tabular}} & \textbf{\begin{tabular}[c]{@{}c@{}}Amputation type\end{tabular}} & \textbf{\begin{tabular}[c]{@{}c@{}}Amputation level and side\end{tabular}} \\ \midrule
\textbf{\textit{Am1}}                                                           & M         & 68           & 50                                                                         & Traumatic                                                          & Transradial (L*)                                                               \\
\textbf{\textit{Am2}}                                                           & M         & NA           & Congenital                                                                 & Congenital                                                         & Transradial (L*)                                                               \\
\textbf{\textit{Am3}}                                                           & M         & 56           & 46                                                                         & Traumatic                                                          & Transradial (R*)                                                               \\
\textbf{\textit{Am4}}                                                           & M         & 30           & 7.5                                                                           & Traumatic                                                          & Wrist disarticulation (L*), shoulder disarticulation (R)                       \\
\textbf{\textit{Am5}}                                                           & M         & 38           & 1.5                                                                        & Traumatic                                                          & Transradial (R*)                                                               \\ \midrule
\multicolumn{6}{c}{(L) and (R) indicate amputation of left or right arm respectively and * indiciates arm used for this study.} \\\bottomrule                                  \end{tabular*}
\end{table}

\begin{table}[!h]
\centering
\caption{Demographics of able-bodied participants}
\label{table:able-demographics}
\begin{tabular}{@{}cccc@{}}
\toprule
\textbf{Subject ID} & \textbf{Age} & \textbf{Sex} & \textbf{Dominant arm} \\ \midrule
\textbf{\textit{Ab1}}                 & 29           & M            & Right                 \\ 
\textbf{\textit{Ab2}}                 & 26           & M            & Right                 \\ 
\textbf{\textit{Ab3}}                 & 28           & M            & Right                  \\ 
\textbf{\textit{Ab4}}                 & 25           & F            & Left                 \\ 
\textbf{\textit{Ab5}}                 & 24           & M            & Right                 \\  \bottomrule	
\end{tabular}
\end{table}

\begin{figure}[!t]
    \centering
    \includegraphics[width=0.75\textwidth]{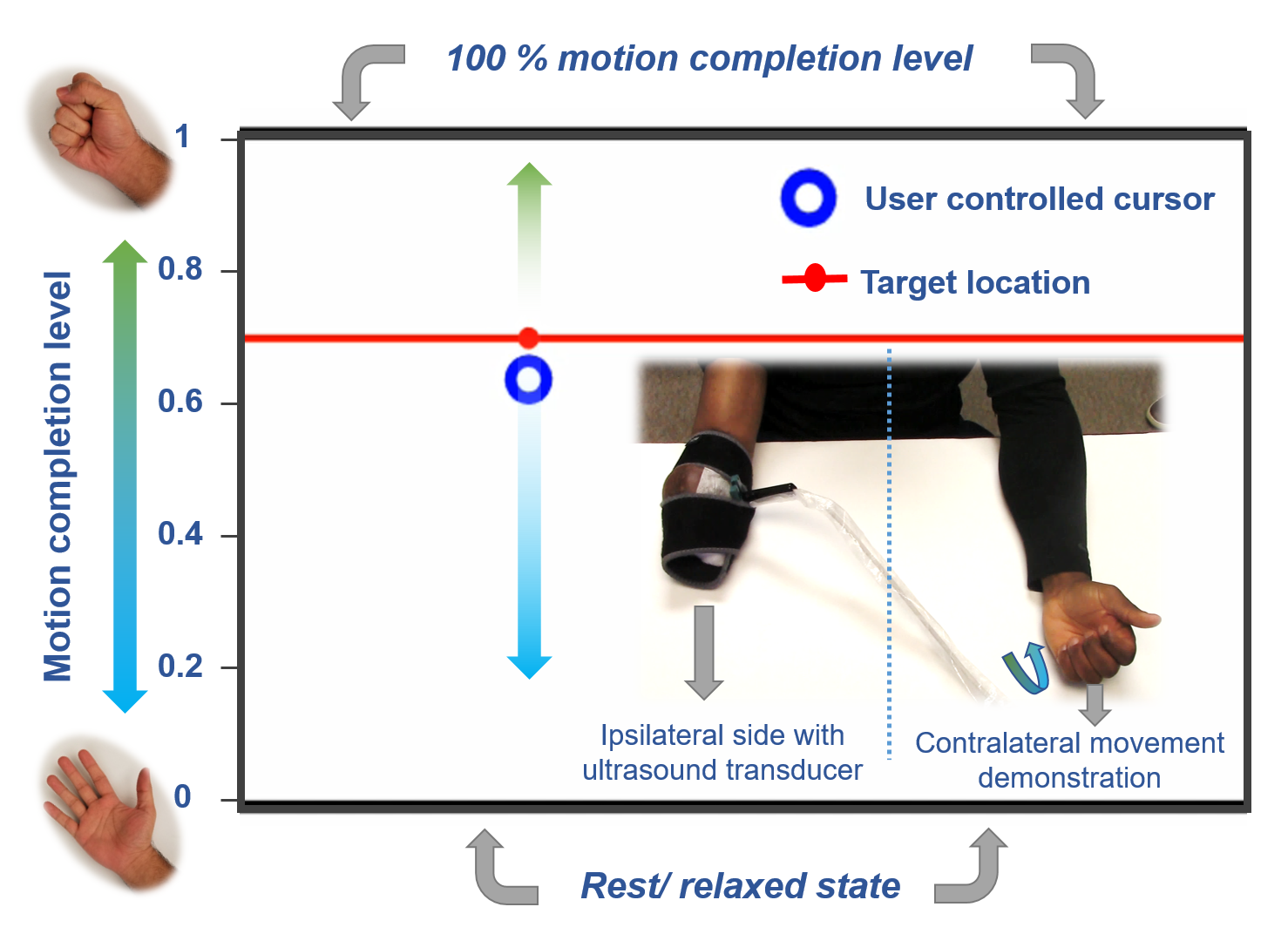}
    \caption{Photo of the experimental setup showing an amputee subject instrumented with an ultrasound transducer on the residuum (inset). The interface for the target holding motion control task described in experiment 2 shows the target position, movement bounds and the cursor position, which is controlled by muscle deformations in the amputee's residuum. Unilateral amputee subjects were asked to demonstrate the perceived motion using their contralateral intact limb.}
    \label{fig:interface-photo}
\end{figure}

For the entire course of this study, all subjects were seated upright with their forearm comfortably supported below the shoulder to minimize fatigue. Subjects were instrumented with a clinical ultrasound system (Terason uSmart 3200T, Terason). The ultrasound system was connected to a low-profile, high-frequency, linear, 16HL7 transducer. For amputee participants, the transducer was positioned on the residuum below the elbow, such that all individual phantom finger movements resulted in considerable movement in the field-of-view. For able-bodied participants, the transducer was manually positioned on the volar aspect of the dominant forearm, approximately 4cm - 5cm from the olecranon process in order to image both the deep and superficial flexor muscle compartments. Additionally, able-bodied participants were asked to place their forearm inside an opaque enclosure that prevented direct observation of arm movements below the elbow. This ensured that able-bodied participants relied solely on their kinesthetic sense to perform all motion control tasks. The transducer was then secured in a custom-designed probe holder and held in place with a stretchable cuff as shown in Fig.\ref{fig:interface-photo} (inset). Ultrasound image sequences from the clinical ultrasound system were acquired using MATLAB (The MathWorks, Inc.) and processed using custom-developed algorithms discussed in the following section. 

\subsection*{Control algorithms} 
\label{sec:methodsAlgo}

Following probe placement, subjects underwent an initial training phase during which they performed repeated iterations of a set of motions (\textit{power grasp, wrist pronation, point, key grasp, tripod}), one motion at a time. Movements were timed to a metronome such that the participant first transitioned from \textit{rest} to the end state of the selected motion, then held that position for a fixed number of metronome beats, then transitioned from the end state back to \textit{rest}, and lastly held for the same number of metronome beats. This process was repeated for 5 repetitions. The first frame of each motion sequence was extracted from the ultrasound data and labeled as '\textit{rest}'. All subsequent frames were compared to the first frame by computing the distance from the \textit{rest} image. This measurement was performed in real-time and a visualization of the signal was provided to participants in real-time as well. An inverse of the computed signal was used for visualization, such that areas of low-similarity to \textit{rest} appeared as peaks and areas of high similarity to \textit{rest} appeared as valleys. The participants were thus able to track the extent to which their muscles were contracted and whether this contraction was consistent over time. Using the same signal, plateaus in similarity were identified in order to find the area where the participant was holding the motion or holding at \textit{rest}. The ultrasound frames at each plateau were averaged into a single representative image and added to a training database with a corresponding motion or \textit{rest} label. The process was repeated until training images corresponding to each movement had been added to the database. The accuracy of the database was then verified by performing leave-one-out cross-validation with a 1-nearest-neighbor classifier using correlation coefficient as a similarity measure. 

\tikzstyle{startstop} = [rectangle, rounded corners, minimum width=1cm, minimum height=1cm,text centered, draw=black, fill=yellow~30]
\tikzstyle{io} = [trapezium, trapezium left angle=70, trapezium right angle=110, minimum width=1cm, minimum height=1cm, text centered, text width=3cm, draw=black, fill=blue!15]
\tikzstyle{process} = [rectangle, minimum width=1cm, minimum height=1cm, text centered, text width=3cm, draw=black, fill=gray!30]
\tikzstyle{decision} = [diamond, minimum width=1cm, minimum height=1cm, text centered, draw=black, fill=green!30]
\tikzstyle{arrow} = [thick,->,>=stealth]

\begin{figure}[!t]
\centering
\begin{tikzpicture}[node distance=1.5cm]
\node (in1) [io] {Incoming image frame at time \textit{t}};
\node (dummy)[right of = in1,xshift=2cm] (dummy) {}; 
\node (in2) [io, right of = in1,xshift=6cm] {Training images for selected motion};
\node (pro3) [process, below of = dummy] {Mean correlation coefficient at time \textit{t}, $c_t$};
\node (pro5) [process, below of = pro3, yshift = -0.3 cm] {Update upper bound ($u$), and lower bound ($l$)};
\node (pro6) [process, below of = pro5,yshift = -0.3cm] {if $l_{t-1} < c_t < u_{t-1}$};
\node (pro7) [process, below of = pro5, xshift=-6.5cm,yshift = -0.3cm]{if $c_t < l_{t-1}$, $l_t = \frac{c_t+l_{t-1}}{2}$};
\node (pro8) [process, below of = pro5, xshift=6.5cm,yshift = -0.3cm]{if $c_t > u_{t-1}$, $u_t = \frac{c_t+u_{t-1}}{2}$};
\node (pro9) [process, below of = pro6, xshift=-4cm] {if $|l_{t-1}-c_t| < |u_{t-1}-c_t|$, $l_t = (0.99)l_{t-1} + (0.01)c_t$};
\node (pro10) [process, below of = pro6, xshift=4cm] {if $|u_{t-1}-c_t| < |l_{t-1}-c_t|$, $u_t = (0.99)u_{t-1} + (0.01)c_t$};
\node (pro11) [io, below of = pro6, yshift = -1.2cm] {Proprioceptive sonomyographic control signal, $p_t = \frac{c_t-l_t}{u_t-l_t}$};

\draw [arrow] (in1) |- (pro3);
\draw [arrow] (in2) |- (pro3);
\draw [arrow] (pro3) -- (pro5);
\draw [arrow] (pro5) -- (pro6);
\draw [arrow] (pro5) -| (pro7);
\draw [arrow] (pro5) -| (pro8);
\draw [arrow] (pro6) |- (pro9);
\draw [arrow] (pro6) |- (pro10);
\draw [arrow] (pro7) |- (pro11);
\draw [arrow] (pro8) |- (pro11);
\draw [arrow] (pro10) |- (pro11);
\draw [arrow] (pro9) |- (pro11);
\end{tikzpicture}
\caption{Image analysis pipeline for computing proportional control signal for an incoming, pre-anotated image frame in real-time. Upper bound ($u$) and lower bound ($l$) are initialized as $\infty$ and $-\infty$ respectively.} \label{fig:propControlFlowchart}
\end{figure}
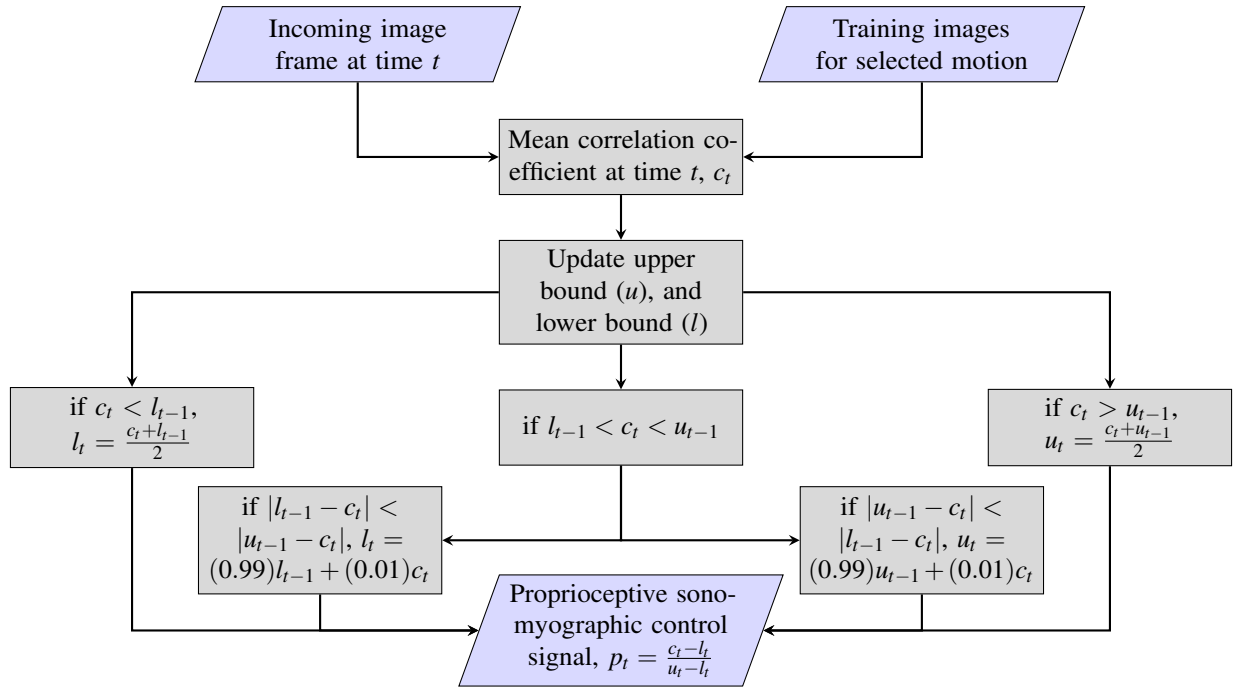
In order to compute motion specific, real-time proprioceptive sonomyographic control signals, a motion was first selected from the training database. The mean correlation value, $c_{t}$, of the acquired image frame at time-point, $t$, versus all frames of the selected motion in the training database was computed and mapped to a motion completion value between '0' and '1'. An upper bound, $u$, and a lower-bound, $l$, were initialized, corresponding to the expected correlation value at the motion end-state and rest respectively so that the correlation signal could be normalized between \textit{rest} ('0') and 100\% motion completion ('1'). Since it is highly unlikely that a participant reaches the exact same motion end state and rest for every single iteration of a given motion, both bounds were dynamically updated as a weighted sum of the instantaneous correlation signal, $c_{t}$ and the closest bound. This process is described in detail in Fig.~\ref{fig:propControlFlowchart}, but the general approach to the bound update procedure is as follows: if a correlation signal that was higher than the expected upper bound was observed, the upper bound was increased; likewise, if a correlation signal lower than the expected lower-bound was observed, the lower bound was decreased; over time, the expected bounds are relaxed very slowly, i.e. the expected upper bound was lowered and the expected lower bound was increased, under the assumption that the bounds are uncertain estimates. 

The effect of the dynamic nature of these updates is that the correlation signal pushes the bounds outwards, resulting in a system that is more responsive if we underestimate the initial bounds than it is if we overestimate. An underestimation would result in the participant being nudged towards 100\% completion or towards rest earlier than expected; whereas an overestimation of the bounds would result in the participant being unable to fully reach 0\% completion (\textit{rest}) or 100\% completion (motion end state). In order to mitigate the system's bias towards \textit{rest} and completion, we performed the bound relaxation at a very slow rate. And though fixed bounds could be used at the risk of overestimation, the dynamic bound adjustment enables the system to adapt in real-time to variations in the participants muscle activation strength due to fatigue, variations in signal amplitude due to sensor movements, or environmental noise and other such factors \cite{Johnson2017}.

\subsection*{Experimental protocols}

\subsubsection*{Experiment 1 - Motion discriminability training}
\label{sec:methodsExp1}
The aim of this experiment was to determine the extent of discriminability that can be achieved across multiple motions for able-bodied and amputee subjects. For this experiment, subjects were asked to perform repetitions of a pre-selected motion, interleaved with \textit{rest} phases between each repetition, in response to audible metronome cues. During the course of the experiment, subjects were provided with a view of ultrasound images acquired from their residuum (or intact limb for able-bodied subjects) in conjunction with the real-time correlation value, $c_{t}$, of the current image frame to \textit{rest} as described in the previous section (also see video in Supplementary material M1).

The study involved blocks of trials, each consisting of five repetitions of a predefined set of motions. Trials were repeated till cross-validation (CV) accuracy exceeded 85\% and subjects reported that they were comfortable performing the motions. All of the amputee subjects listed in Table \ref{table:amp-demographics} participated in this experiment. Subject-specific motion sets and number of iterations performed by each amputee participant are listed in Table \ref{table:sub-uims}. All able-bodied participants performed five iterations of \textit{power grasp}, \textit{wrist pronation}, \textit{tripod}, \textit{key grasp}, and \textit{point} each. Outcome measures for this experiment were leave-one out cross-validation accuracies for the first trial and the best trial performed, as well as the average accuracy over all trials. 

\begin{table}[!h]
\centering
\caption{User-intended motions and number of iteration of each motion performed by amputee subjects.}
\label{table:sub-uims}
\begin{tabular*}{\textwidth}{P{0.15\textwidth}P{0.4\textwidth}P{0.4\textwidth}}
\toprule
\textbf{Subject ID}          & \textbf{Motions performed}          & \textbf{Number of iterations per motion}         \\
\hline
\textbf{\textit{Am1}}                & PG, WP, Tr, KG, Po                  & 20                                      \\
\textbf{\textit{Am2}}                 & PG, WP, In, Tr                       & 5                                       \\
\textbf{\textit{Am3}}*                 & PG, WP, Tr, KG, Po, In, Tr           & 5 (S1), 25 (S2)                         \\
\textbf{\textit{Am4}}                 & PG, WP, Tr, KG, Po                  & 20                                      \\
\textbf{\textit{Am5}}*                  & PG, WP, Tr, KG, Po                  & 15 (S1), 25 (S2)                        \\
\hline
\multicolumn{3}{c}{PG = \textit{power grasp}, WP = \textit{wrist pronation}, Tr = \textit{tripod}, KG = \textit{key grasp}, Po = \textit{point}, In = \textit{index flexion}}\\
\hline
\multicolumn{3}{c}{* Motions performed over two different sessions, where S1 = session 1 and  S2 = session 2.}                                   \\\bottomrule  
\end{tabular*}
\end{table}

\subsubsection*{Experiment 2 - Proportional target holding task}\label{sec:exp2}
\label{sec:methodsExp2}

The aim of this experiment was to quantify \textit{proprioceptive sonomyographic control} performance of amputees and able-bodied subjects at graded muscle activation levels for multiple motions. A motor control task was implemented where the participant controlled an on-screen cursor that could move up or down in proportion to the degree of muscle activation in the forearm as shown in Fig.~\ref{fig:interface-photo}. The cursor on the computer screen could move up towards a normalized bound of '1' in proportion to the performed completion level of a selected motion, reaching '1' when the motion was at 100\% completion. Similarly, the cursor could move down towards the normalized bound of '0' as the user returned from motion completion towards \textit{rest}, reaching '0' when the user was completely at \textit{rest} (see video in supplementary material M2).

After an initial calibration step to initialize the bounds, the control interface presented the user with a target position randomly chosen from a predefined set of quantized, equidistant positions ($N_{P}$), between the normalized upper and lower bounds. The target remained fixed at that position for a set hold-time, $T_{H}$ and then moved to the next position until all points were exhausted. For each target position, the participant was prompted to move the cursor to the target by contracting or relaxing their muscles and holding the cursor position until the hold period expired. Unilateral amputee subjects were also asked to demonstrate the perceived motion of their phantom limb using their intact, contralateral arm. Fig.~\ref{fig:interface-photo} shows a screen-shot of the control interface with the target and the user-controlled cursor.  

We first conducted a pilot study with \textit{Am2} and \textit{Am3}, with five quantized positions ($N_{P}=5$) and a hold time, $T_{H} =15$s to validate our control algorithms. For the pilot study, the hold period commenced when the cursor entered the current target’s quantization bounds, defined as $Q=\pm100/[2\times(N_{P}-1)]\%$ around the target. If the user failed to enter the quantization bounds within a timeout period, $T_{to}=30$s, the target automatically moved to the next set-point.

Following the pilot study, all able-bodied participants and amputee subjects \textit{Am3}, \textit{Am4} and \textit{Am5} were recruited to perform the same motion control task with eleven graded levels ($N_{P}=11$) and hold time of $T_{H} =10$s. For this extended study, the hold period commenced as soon as the target was presented to the user, irrespective of whether the user was able to reach the target. At the end of the hold period, the target moved to the next target location, randomly chosen from the set of eleven target positions, without replacement. All participants performed three trials of each target level, for a total of 33 target positions. 

Position error, stability error, task completion rate and movement time served as evaluation metrics for \textit{proprioceptive sonomyographic control} performance. Position error was computed as the mean error between the cursor and the target position, while stability error was calculated to be the standard deviation of the cursor position from the target position. Both, position and stability errors were computed for the time when the cursor first entered the quantization bound $Q$ till the hold time, $T_{H}$ expired. Task completion rate was defined as the percentage of targets that the user was able to successfully reach of the total target locations presented. A target was considered to have been successfully acquired if the cursor entered the quantization bound of $Q$ around the target position. However, the user was not required to stay within the bound, $Q$, for successful target acquisition. The time starting from when the target was first presented to the user, to when the cursor first entered the quantization bounds $Q$ was measured to be the movement time.


\begin{figure}[!h]
    \centering
\includegraphics[width=\textwidth]{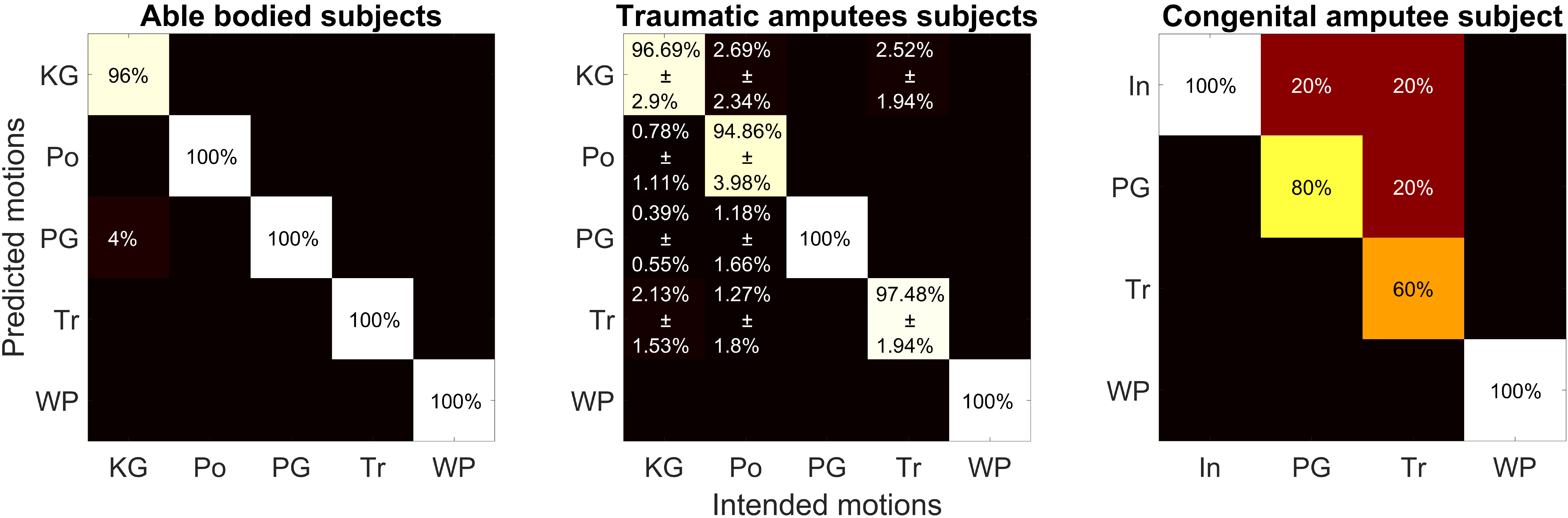}\vspace*{1em}
\includegraphics[width=\textwidth]{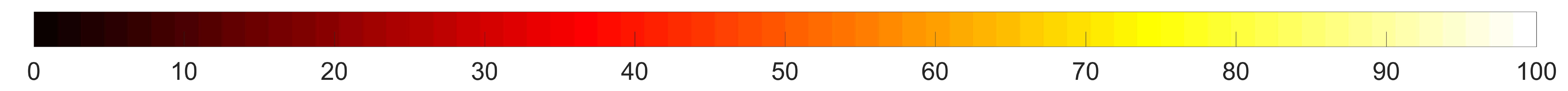}\\
\includegraphics[width=0.6\textwidth]{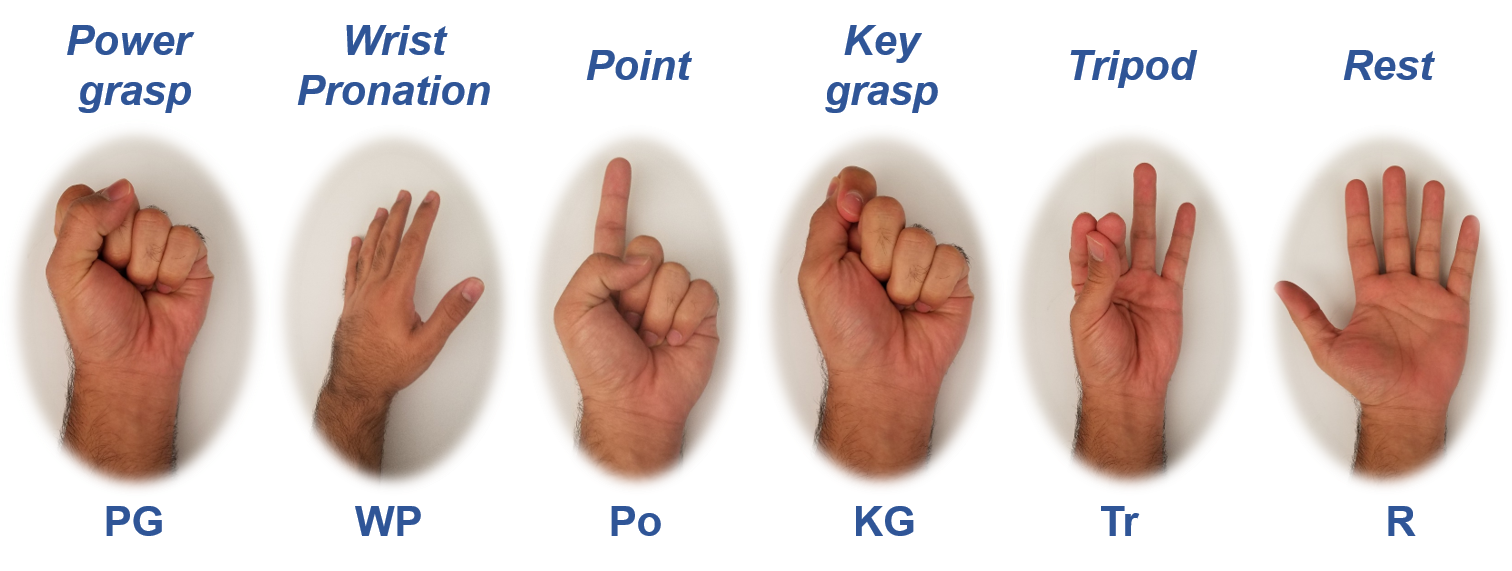} 
    \caption{Aggregate confusion matrices showing post-training, motion discriminability for able-bodied, traumatic and congenital amputee subjects. Motion legend- PG = \textit{power grasp}, WP = \textit{wrist pronation}, Po = \textit{point}, KG = \textit{key grasp}, Tr = \textit{tripod}, In = \textit{index flexion}.}
    \label{fig:confMat}
\end{figure}

\section*{Results}
\subsection*{Experiment 1}

The 1-nearest neighbor classifier was first validated with able-bodied subjects. Four out of the five able-bodied participants achieved cross-validation accuracies of 100\% within five repetitions (or one trial) across all five motions. Fig.~\ref{fig:confMat} shows the aggregate confusion matrix for all able-bodied subjects. It shows that four out of five motions were predicted with 100\% accuracy and \textit{key grasp} was incorrectly predicted as \textit{power grasp} in just one out of 25 motion instances (5 subjects performing 5 motions each).  

All of the amputee subjects, including the congenital amputee (\textit{Am2}), were also able to successfully complete the tasks, with an average prediction accuracy of 96.8$\pm$5.4\% for at least 4 motions. Fig.~\ref{fig:confMat} shows the aggregate confusion matrix for all traumatic amputees and the congenital amputee subject, \textit{Am2}. For traumatic amputees, the average cross-validation accuracy was 96.76\% for five motions, with \textit{key grasp} and \textit{point} having the lowest prediction accuracies. In contrast, the congenital amputee subject achieved a cross-validation accuracy of 85\% for four motions with \textit{tripod} having the least prediction accuracy at 60\%. The subject reported not having phantom sensation of a thumb, which could have negatively affected motion discriminability for motions involving thumb movements, such as \textit{tripod} and \textit{power grasp}. All subjects typically completed the training phase in an hour or less.

\begin{table}[!b]
\centering
\caption{Comparison of cross-validation accuracies for amputee subjects with and without \textit{rest} as a motion class.}
\label{table:ampCV}
\resizebox{\textwidth}{!}{%
\begin{tabular}{@{}c|ccc|ccc@{}}
\toprule
\multirow{3}{*}{\textbf{Subject ID}} & \multicolumn{6}{c}{\textbf{Cross-validation accuracy (\%)}}                                                                                                                                                                                             \\  
                                     & \multicolumn{3}{c}{\textbf{Including rest}}                                                                                & \multicolumn{3}{c}{\textbf{Excluding rest}}                                                                                \\\cmidrule(l){2-7}
                                     & \textbf{First trial} & \textbf{Best trial} & \textbf{\begin{tabular}[c]{@{}c@{}}Average across \\ all trials\end{tabular}} & \textbf{First trial} & \textbf{Best trial} & \textbf{\begin{tabular}[c]{@{}c@{}}Average across \\ all trials\end{tabular}} \\ \cmidrule(r){1-7}
\textit{\textbf{Am1}}                & 88.33                & 90.83               & 87.71                                                                         & 100.00               & 100.00              & 99.00                                                                         \\
\textit{\textbf{Am2}}*                & 80.50                & 80.50               & 80.50                                                                         & 85.00                & 85.00               & 85.00                                                                         \\
\textit{\textbf{Am3 (S1)}}*           & 94.12                & 94.12               & 94.12                                                                         & 100.00               & 100.00              & 100.00                                                                        \\
\textit{\textbf{Am3 (S2)}}           & 82.50                & 95.00               & 86.33                                                                         & 96.00                & 100.00              & 96.80                                                                         \\
\textit{\textbf{Am4}}                & 100.00               & 100.00              & 100.00                                                                        & 100.00               & 100.00              & 100.00                                                                        \\
\textit{\textbf{Am5 (S1)}}           & 86.67                & 86.67               & 85.00                                                                         & 100.00               & 100.00              & 100.00                                                                        \\
\textit{\textbf{Am5 (S2)}}           & 92.50                & 93.33               & 89.33                                                                         & 96.00                & 100.00              & 96.80                                                                         \\\midrule
\textbf{Mean $\pm$ SD}               & \textbf{89.23$\pm$6.82}       & \textbf{91.49$\pm$6.32}      & \textbf{89.00$\pm$6.38}               & \textbf{96.71$\pm$5.50} & \textbf{97.86$\pm$5.67} & \textbf{96.80$\pm$5.40}                                                                         \\ \midrule
\multicolumn{7}{c}{*Subjects performed one trial of five repetitions. Refer to Table \ref{table:sub-uims}}                                                                                                                                                                                                                           \\ \bottomrule
\end{tabular}
}
\end{table}

We also analyzed the influence of treating \textit{rest} as a separate motion class on the prediction accuracies. When \textit{rest} was excluded from cross-validation, motion discriminability improved for all participants. For able-bodied subjects, the five motions were predicted correctly with 99.2\% cross-validation accuracy. For amputee subjects, the mean cross-validation accuracy for all trials increased by 7.8\% to 96.8\% when \textit{rest} was excluded as shown in Table~\ref{table:ampCV}. 

\begin{figure*}[ht!]
  \begin{subfigure}[t]{\linewidth}
    \centering
    \includegraphics[width = 0.7\textwidth]{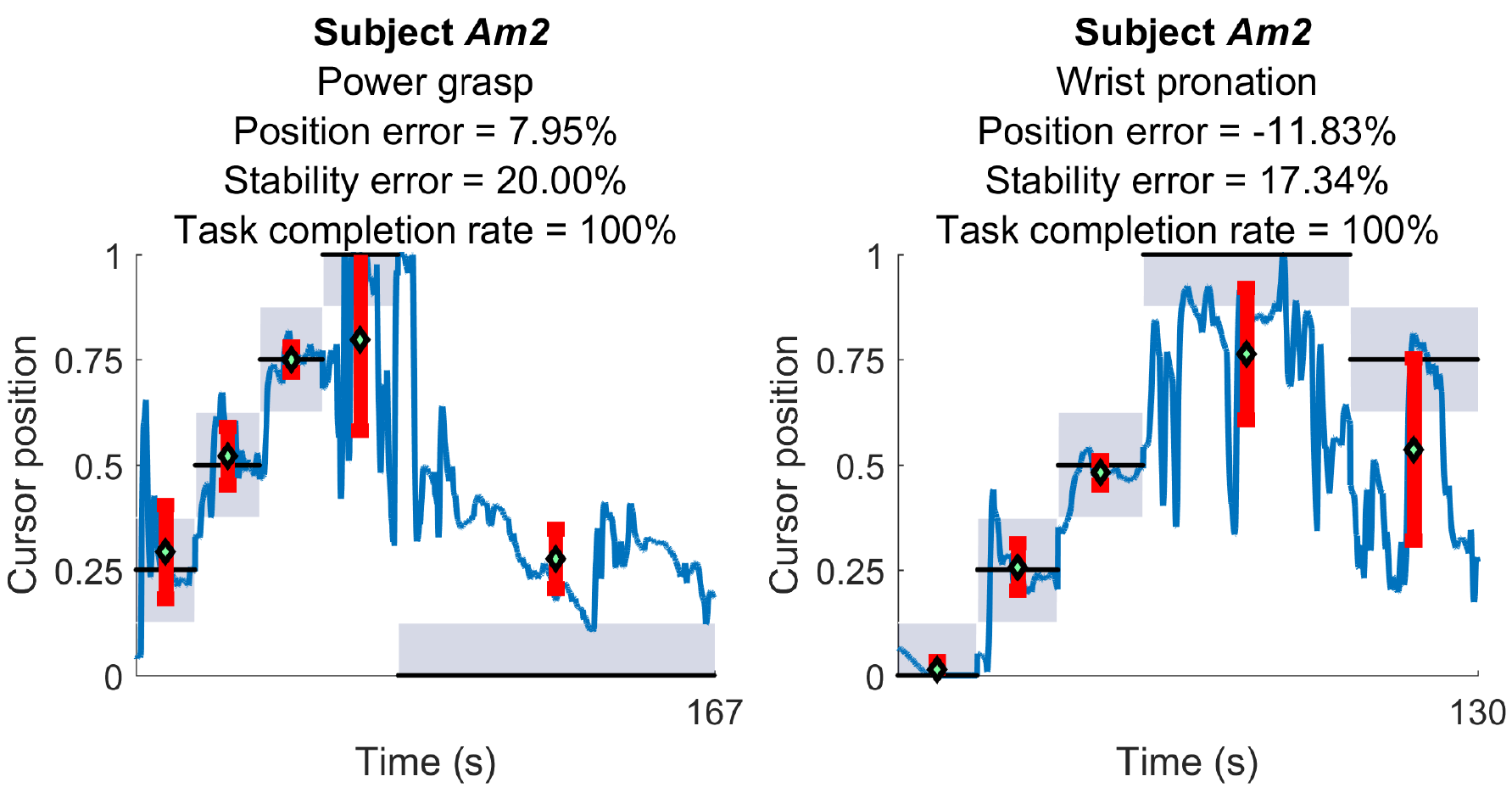}
    \caption{}\label{fig:pilotTH1}
  \end{subfigure}
  \begin{subfigure}[t]{\linewidth}
    \centering
    \includegraphics[width = 0.7\textwidth]{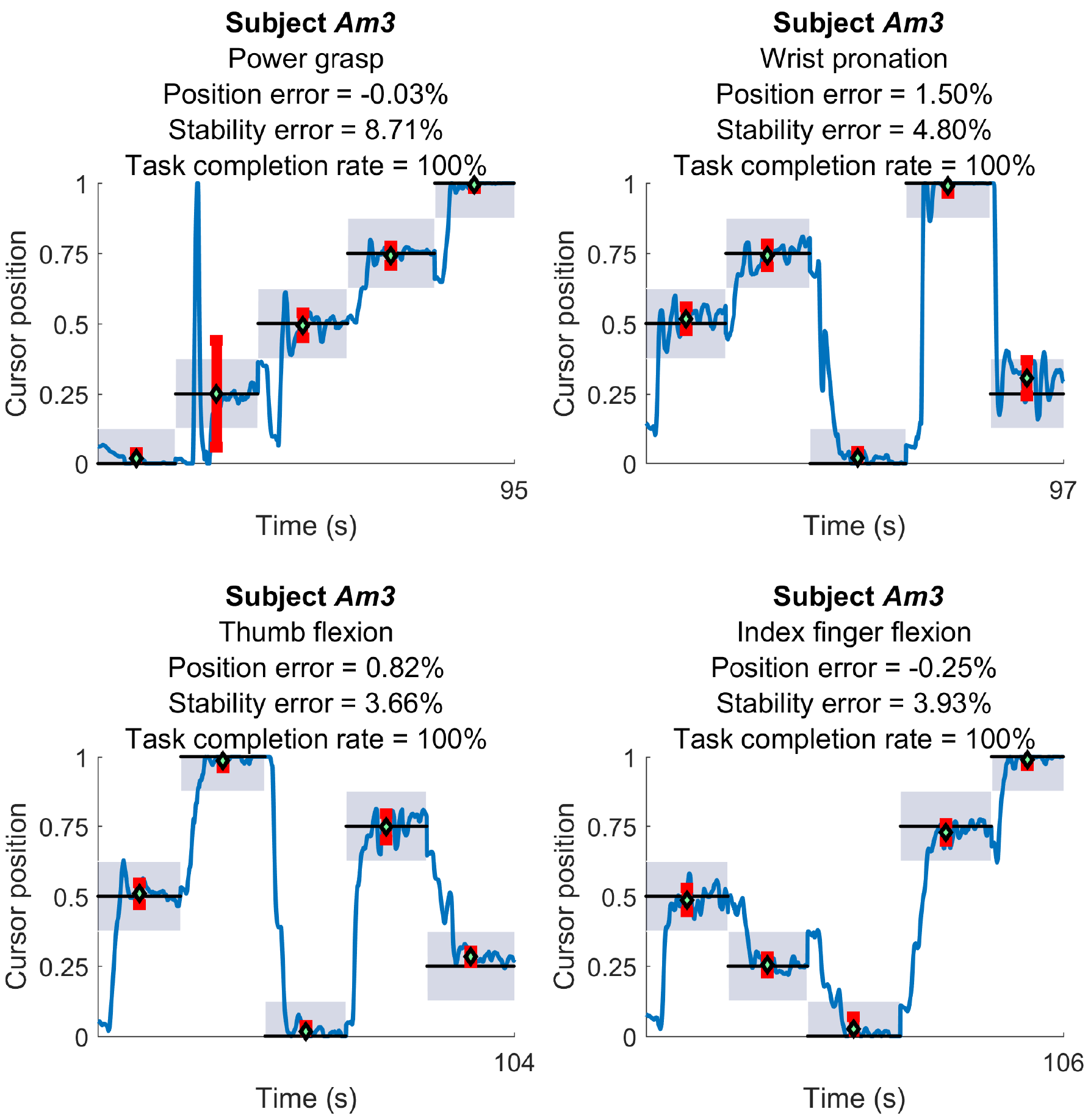}
    \caption{}\label{fig:pilotTH2}
  \end{subfigure}  
  \caption{Plots of user-controlled cursor position against target position for the pilot target holding task, for \textit{Am2} (\ref{fig:pilotTH1}) and \textit{Am3} (\ref{fig:pilotTH2}). The target randomly moved between five quantized target levels within normalized bounds of \textit{rest} (\lq0\rq) and motion completion (\lq1\rq). Position and stability errors for individual target position segments are also shown.}
  \label{fig:pilotTH}
\end{figure*}

Table~\ref{table:ampCV} also shows cross-validation accuracies with and without \textit{rest} for the initial and best set of trials for amputee subjects. In both cases, the mean cross-validation accuracy for the initial trial was comparable to best accuracy figures showing that the classifier was able to consistently predict motions when presented with representative image frames repeated across several trials.

\begin{figure*}[ht!]
  \begin{subfigure}[c]{1\linewidth} \centering
    \includegraphics[width=0.3\linewidth]{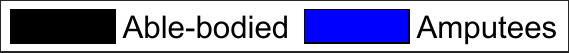}
  \label{fig:aggTHLegend}
  \end{subfigure}
  \centering \begin{subfigure}[t]{0.4\linewidth}
    \includegraphics[width=1\linewidth]{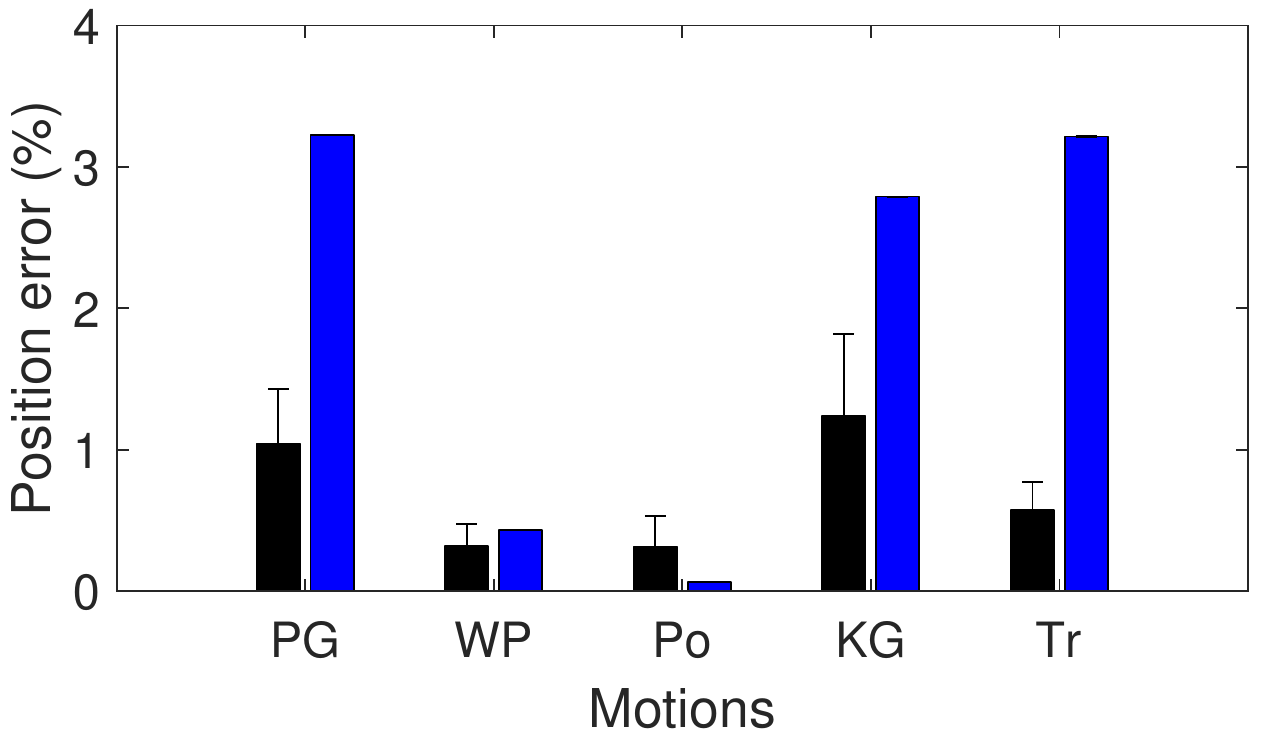}
    \caption{}\label{fig:posErr}
  \end{subfigure}
  \begin{subfigure}[t]{0.4\linewidth}\centering
    \includegraphics[width=1\linewidth]{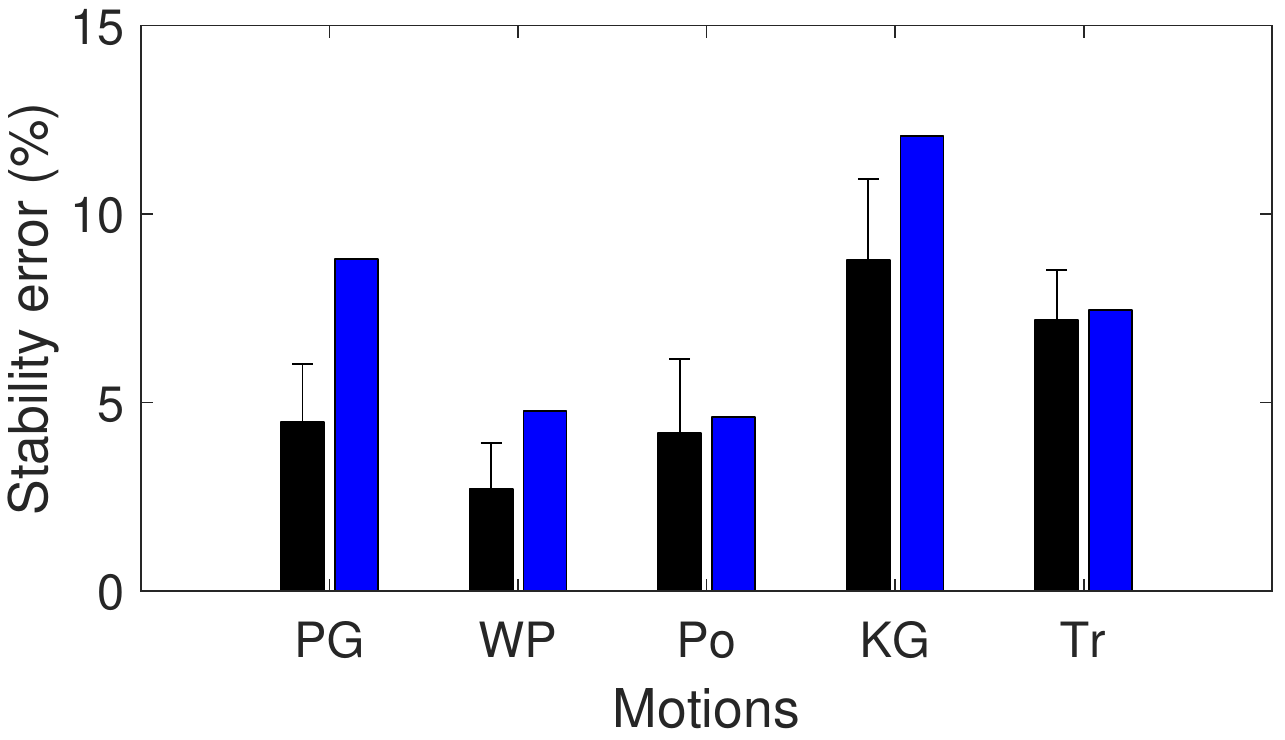}
    \caption{}\label{fig:stabErr}
  \end{subfigure}  
  \begin{subfigure}[t]{0.4\linewidth}\centering
    \includegraphics[width=1\linewidth]{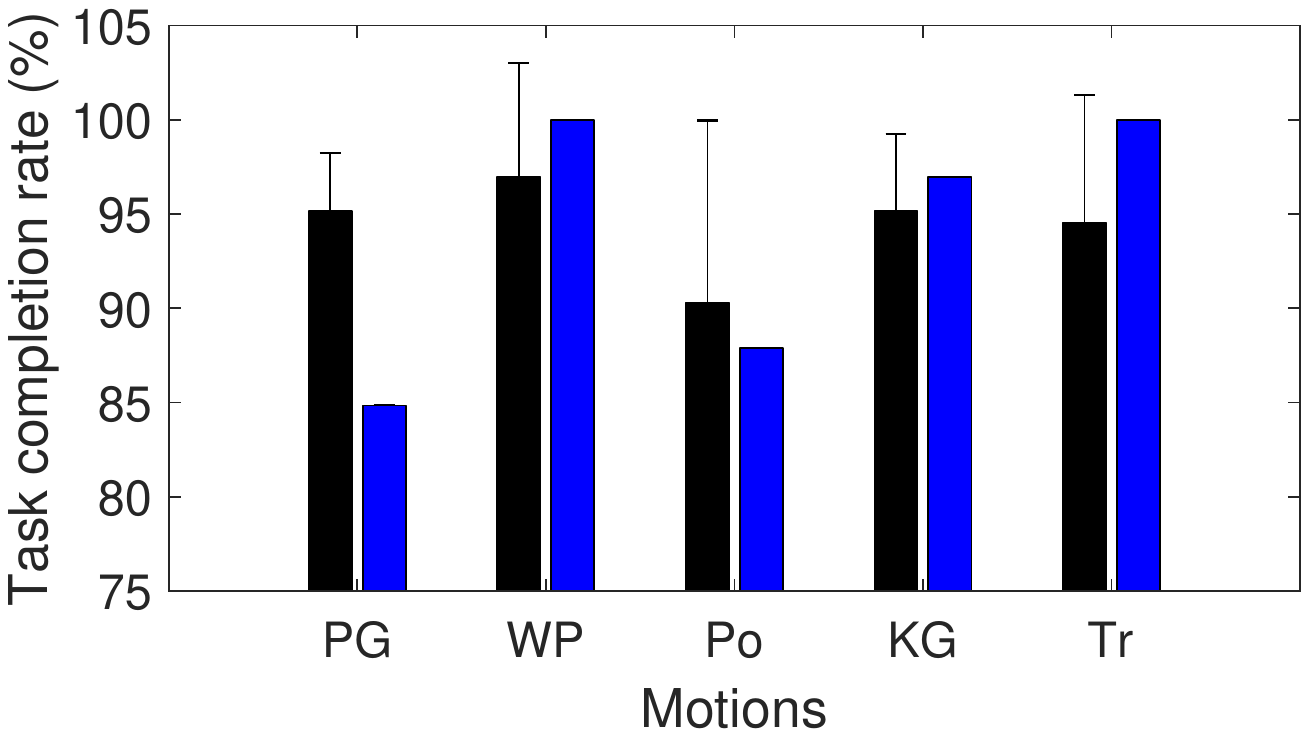}
    \caption{}\label{fig:compRateMotion}
  \end{subfigure}
  \begin{subfigure}[t]{0.4\linewidth}\centering
    \includegraphics[width=1\linewidth]{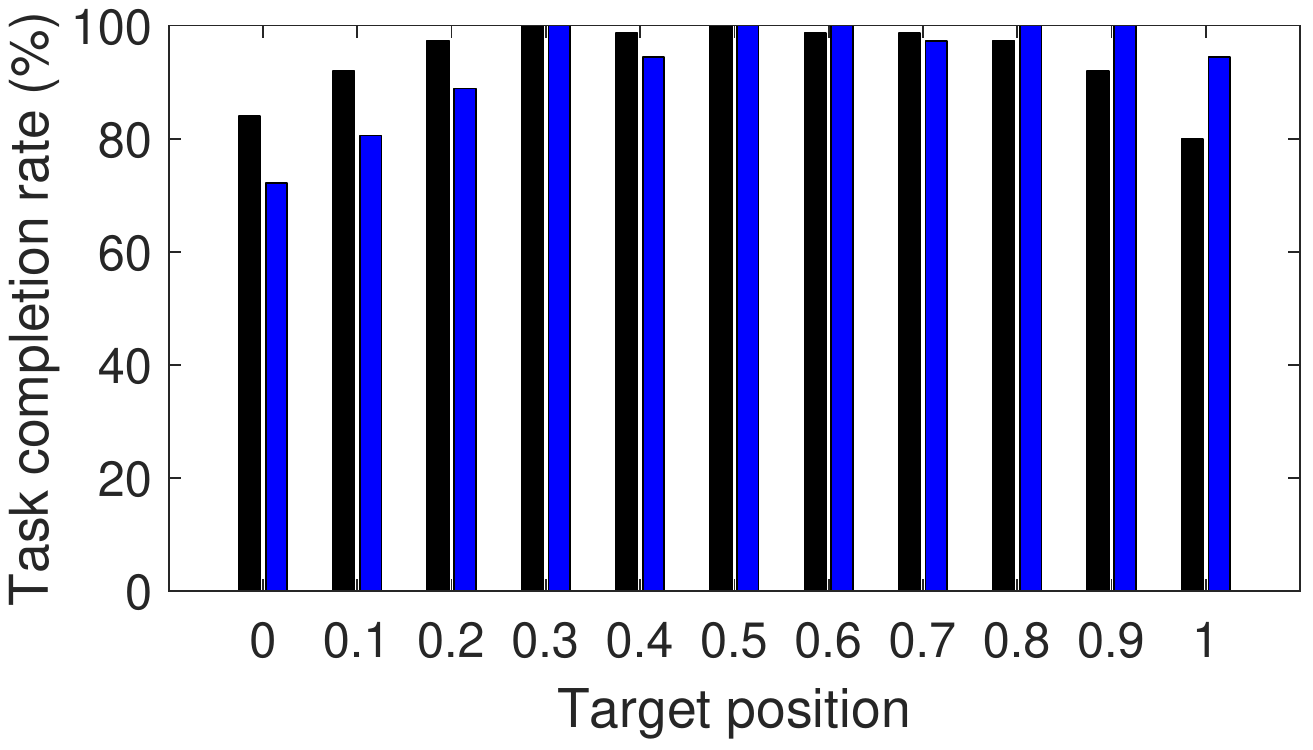}
    \caption{}\label{fig:compRateTargetPos}
  \end{subfigure}
    \begin{subfigure}[c]{1\linewidth} \centering
    \includegraphics[width=0.5\linewidth]{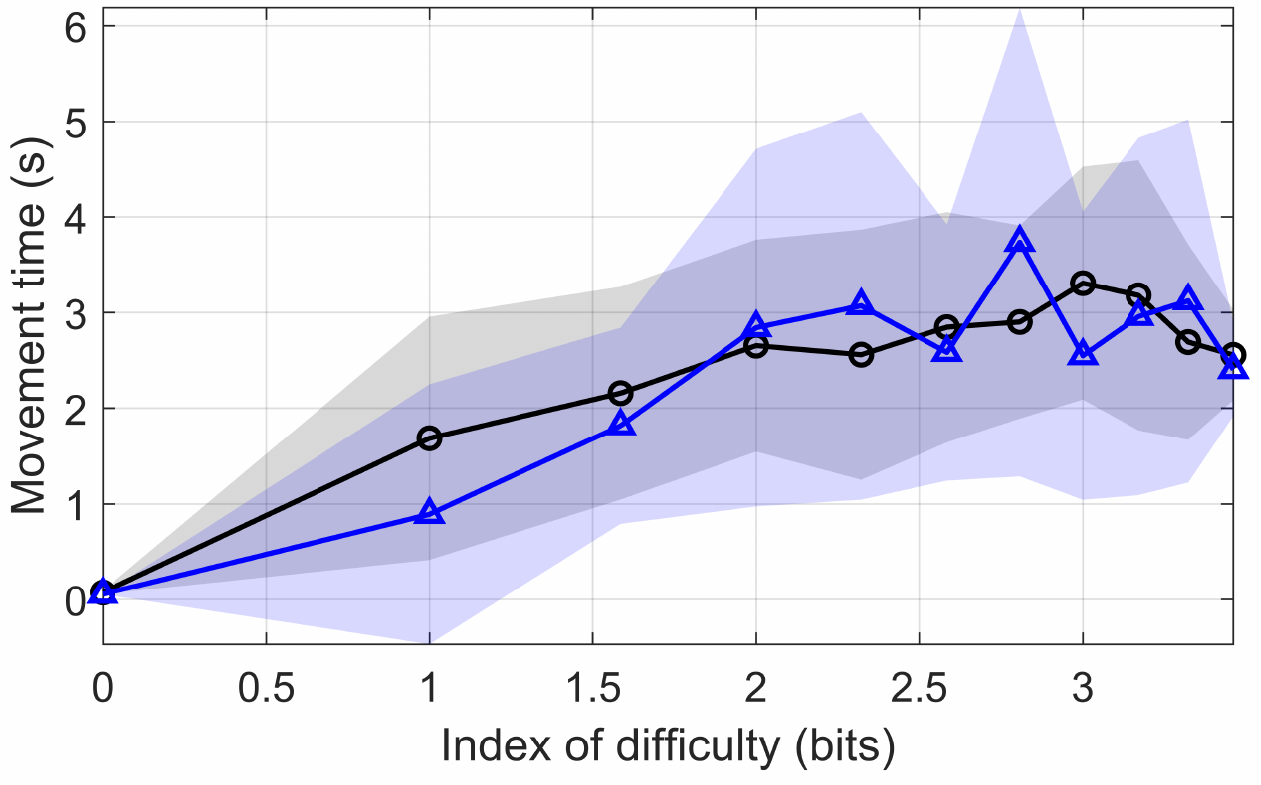}
  \caption{}\label{fig:movementTime}
  \end{subfigure}

  \caption{Aggregate outcome metrics and Fitt's law analysis for the target holding task (experiment 2) for able-bodied and amputee subjects. Motion legend- PG = \textit{power grasp}, WP = \textit{wrist pronation}, Po = \textit{point}, KG = \textit{key grasp}, Tr = \textit{tripod}.}
  \label{fig:aggregateStatsTH}
\end{figure*}

\subsection*{Experiment 2}
Figs.~\ref{fig:pilotTH1} and \ref{fig:pilotTH2} show results of the pilot target holding study of experiment 2 for amputee subjects, \textit{Am2} and \textit{Am3} respectively. The quantized target trajectory, user-controlled cursor position and associated quantization bounds have been plotted against time. Both subjects were able to successfully reach all of the targets presented in the pilot study without any prior training on performing the task. Congenital amputee subject, \textit{Am2} performed \textit{power grasp} with position and stability error of 7.95\% and 20\% respectively. For \textit{wrist pronation} position error was 11.83\% whereas stability error was 17.83\%. Amputee subject \textit{Am3}'s position errors ranged between -0.03\% to 1.50\% while stability error ranged from 3.66\% to 8.71\% across four motions. \textit{Power grasp} had the lowest position error whereas \textit{thumb flexion} was found to have the lowest stability error.

Table~\ref{table:ableTHmetrics} and Table \ref{table:ampTHmetrics} show outcome metrics for the subsequent, extended target holding holding task performed at eleven graded target levels for able-bodied and amputee subjects respectively. All of the able-bodied participants and amputee subjects, \textit{Am4} and \textit{Am5} performed five motions while subject \textit{Am3} performed 2 motions. \textit{Am3} also performed \textit{key grasp}, however, that data was excluded from analysis due to error during the bound calibration stage. 

\begin{table}[!h]
\centering
\caption{Evaluation metrics for target holding task (experiment 2) for able-bodied subjects.}
\label{table:ableTHmetrics}
\resizebox{\textwidth}{!}{%
\begin{tabular}{@{}c|ccccc|ccccc|ccccc@{}}
\toprule
\multirow{2}{*}{\textbf{\begin{tabular}[c]{@{}c@{}}Subject \\ ID\end{tabular}}} & \multicolumn{5}{c}{\textbf{Position error (\%)}}                    & \multicolumn{5}{c}{\textbf{Stability error (\%)}}                   & \multicolumn{5}{c}{\textbf{ Task completion rate (\%)}}                   \\ \cmidrule(l){2-6} \cmidrule(l){7-11} \cmidrule(l){12-16}
                                                                                & \textbf{PG} & \textbf{WP} & \textbf{Po} & \textbf{KG} & \textbf{Tr} & \textbf{PG} & \textbf{WP} & \textbf{Po} & \textbf{KG} & \textbf{Tr} & \textbf{PG} & \textbf{WP} & \textbf{Po} & \textbf{KG} & \textbf{Tr} \\ \midrule
\textit{\textbf{Ab1}}                & -0.74       & 0.47        & 0.67        & 0.63        & -0.41       & 3.66        & 1.88        & 4.02        & 8.45        & 5.55        & 93.94       & 100.0      & 100.0       & 87.88       & 96.97       \\
\textit{\textbf{Ab2}}                & 1.17        & 0.41        & -0.44       & 1.21        & -0.58       & 4.90        & 3.97        & 1.03        & 8.87        & 7.92        & 90.91       & 100.0      & 96.97       & 96.97       & 100.0      \\
\textit{\textbf{Ab3}}                & -0.66       & 0.04        & 0.22        & -0.69       & -0.38       & 2.98        & 1.66        & 3.52        & 5.18        & 9.17        & 93.94       & 100.0      & 81.82       & 100.0       & 100.0      \\
\textit{\textbf{Ab4}}                & -0.92       & -0.25       & 0.16        & 1.43        & -0.93       & 3.62        & 1.62        & 5.63        & 9.63        & 5.97        & 96.97       & 100.0      & 75.76       & 96.97       & 81.82       \\
\textit{\textbf{Ab5}}                & -1.73       & -0.42       & 0.09        & -2.23       & -0.57       & 7.30        & 4.39        & 6.78        & 11.8        & 7.38        & 100.0       & 84.85      & 96.97       & 93.94       & 93.94       \\
\midrule\textbf{Average}                     & 1.04       & 0.32        & 0.32       & 1.24        & 0.57       & 4.49        & 2.70        & 4.20        & 8.79        & 7.20        & 95.15       & 96.97      & 90.30       & 95.15       & 94.55      \\\midrule
\multicolumn{16}{c}{PG = \textit{power grasp}, WP = \textit{wrist pronation}, Po = \textit{point}, KG = \textit{key grasp}, Tr = \textit{tripod}} \\\bottomrule
\end{tabular}%
}
\end{table}

\begin{table}[!h]
\centering
\caption{Evaluation metrics for target holding task (experiment 2) for amputee subjects.}
\label{table:ampTHmetrics}
\resizebox{\textwidth}{!}{%
\begin{tabular}{@{}c|ccccc|ccccc|ccccc@{}}
\toprule
\multirow{2}{*}{\textbf{\begin{tabular}[c]{@{}c@{}}Subject \\ ID\end{tabular}}} & \multicolumn{5}{c}{\textbf{Position error (\%)}}                    & \multicolumn{5}{c}{\textbf{Stability error (\%)}}                   & \multicolumn{5}{c}{\textbf{Task completion rate (\%)}}                   \\ \cmidrule(l){2-6} \cmidrule(l){7-11} \cmidrule(l){12-16}
                                                                                & \textbf{PG} & \textbf{WP} & \textbf{Po} & \textbf{KG} & \textbf{Tr} & \textbf{PG} & \textbf{WP} & \textbf{Po} & \textbf{KG} & \textbf{Tr} & \textbf{PG} & \textbf{WP} & \textbf{Po} & \textbf{KG} & \textbf{Tr} \\ \midrule
\textit{\textbf{Am3}}                                                           & 2.37        & -0.23       & -           & -           & -           & 10.19       & 4.61        & -           & -           & -           & 69.70       & 100.0      & -           & -           & -           \\
\textit{\textbf{Am4}}                                                           & -0.39       & -0.25       & 0.12        & -0.49       & -0.02       & 2.33        & 5.79        & 3.37        & 8.55        & 4.00        & 96.97       & 100.0      & 96.97       & 96.97       & 100.0      \\
\textit{\textbf{Am5}}                                                           & -6.91       & -0.82       & -0.01       & -5.09       & -6.41       & 13.89       & 3.92        & 5.86        & 15.60       & 10.90       & 87.88       & 100.0      & 78.79       & 96.97       & 100.0      \\
\midrule \textbf{Average}                                                                & 3.22       & 0.43       & 0.07        & 2.79       & 3.22       & 8.80        & 4.77        & 4.62        & 12.08       & 7.45        & 84.85       & 100.0      & 87.88       & 96.97       & 100.0      \\ \midrule
\multicolumn{16}{c}{PG = \textit{power grasp}, WP = \textit{wrist pronation}, Tr = \textit{tripod}, KG = \textit{key grasp}, Po = \textit{point}} \\\bottomrule
\end{tabular}%
}
\end{table}

The average outcome metrics for the target holding task are also shown in Fig.~\ref{fig:aggregateStatsTH}. For the targets that were successfully acquired, Figs.~\ref{fig:posErr} and \ref{fig:stabErr} show that able-bodied subjects performed all motions except \textit{point} with lower position and stability errors compared to amputee subjects. Although, stability errors for able-bodied subjects were lower than amputee subjects, there seems to be a correspondence between motion-specific stability errors across subjects. Motions with high stability errors in able-bodied subjects also correspondingly have high stability errors in amputees. 

The type of motion also seems to be have an influence on position errors and stability errors for all subjects, i.e. motions with higher position error also have high stability error and vice versa. For example, as shown in Figs.~\ref{fig:posErr} and \ref{fig:stabErr}, for able-bodied subjects, \textit{key grasp} followed by \textit{power grasp} and \textit{tripod} have the highest stability and position errors. Similarly, for amputee subjects, a similar trend is observed wherein \textit{key grasp}, \textit{power grasp} and \textit{tripod} exhibited the the highest stability and position errors out of the 5 motions that were performed.

\begin{figure}[!ht]
\centering
\includegraphics[width = \textwidth]{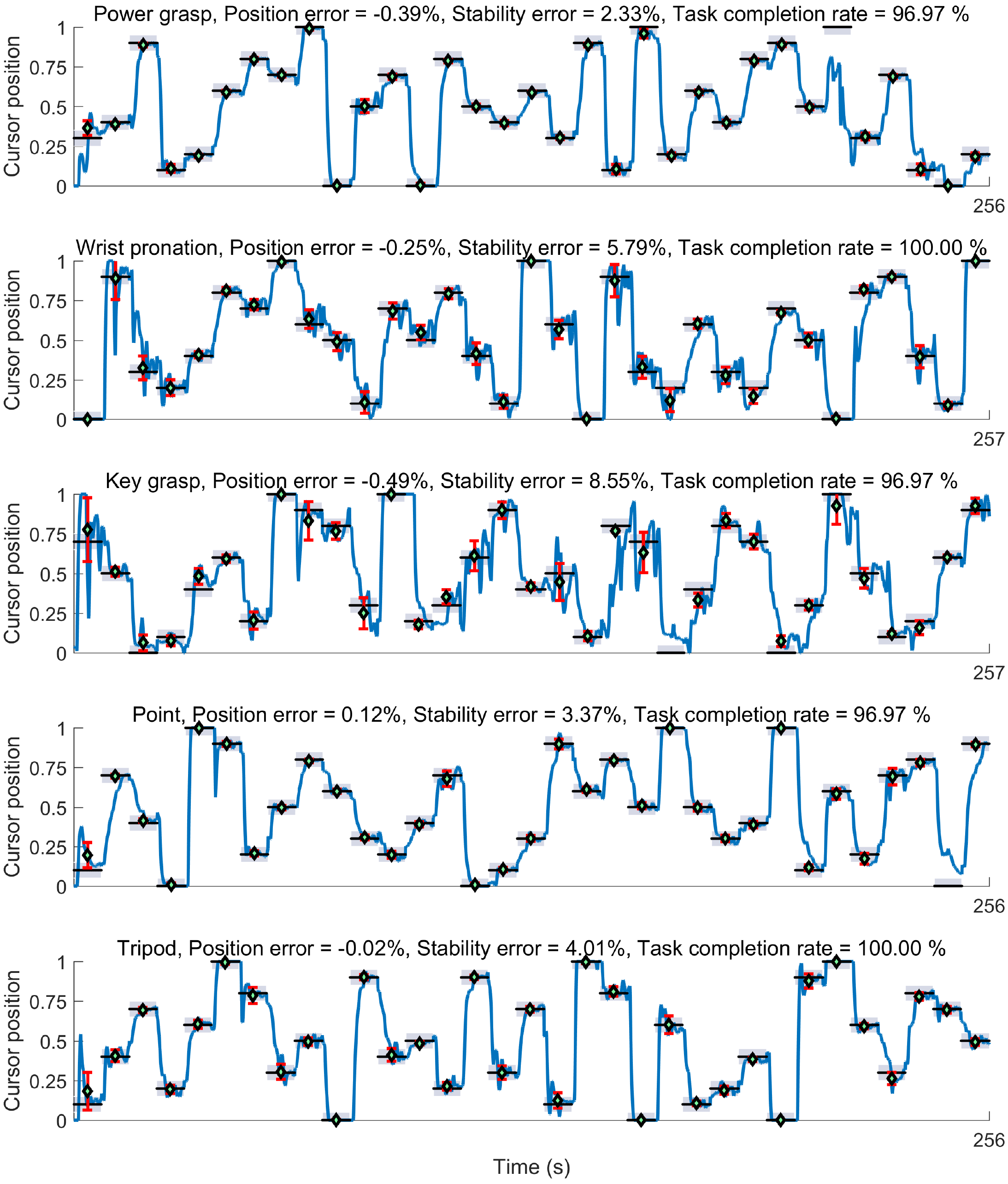}\\
\includegraphics[width=0.6\textwidth]{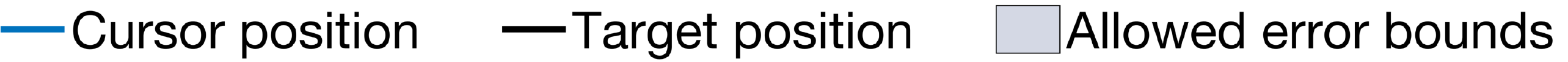}
\label{fig:am4TH}
\caption{Timeseries plots showing amputee subject \textit{Am4}'s performance in the target holding task. The user-controlled cursor position and target locations for five motions along with position and stability errors are shown. Motion legend- PG = \textit{power grasp}, WP = \textit{wrist pronation}, Po = \textit{point}, KG = \textit{key grasp}, Tr = \textit{tripod}.}
\label{fig:Am4TargetHolding}
\end{figure}

\begin{figure}[!ht]
\centering
\includegraphics[width = \textwidth]{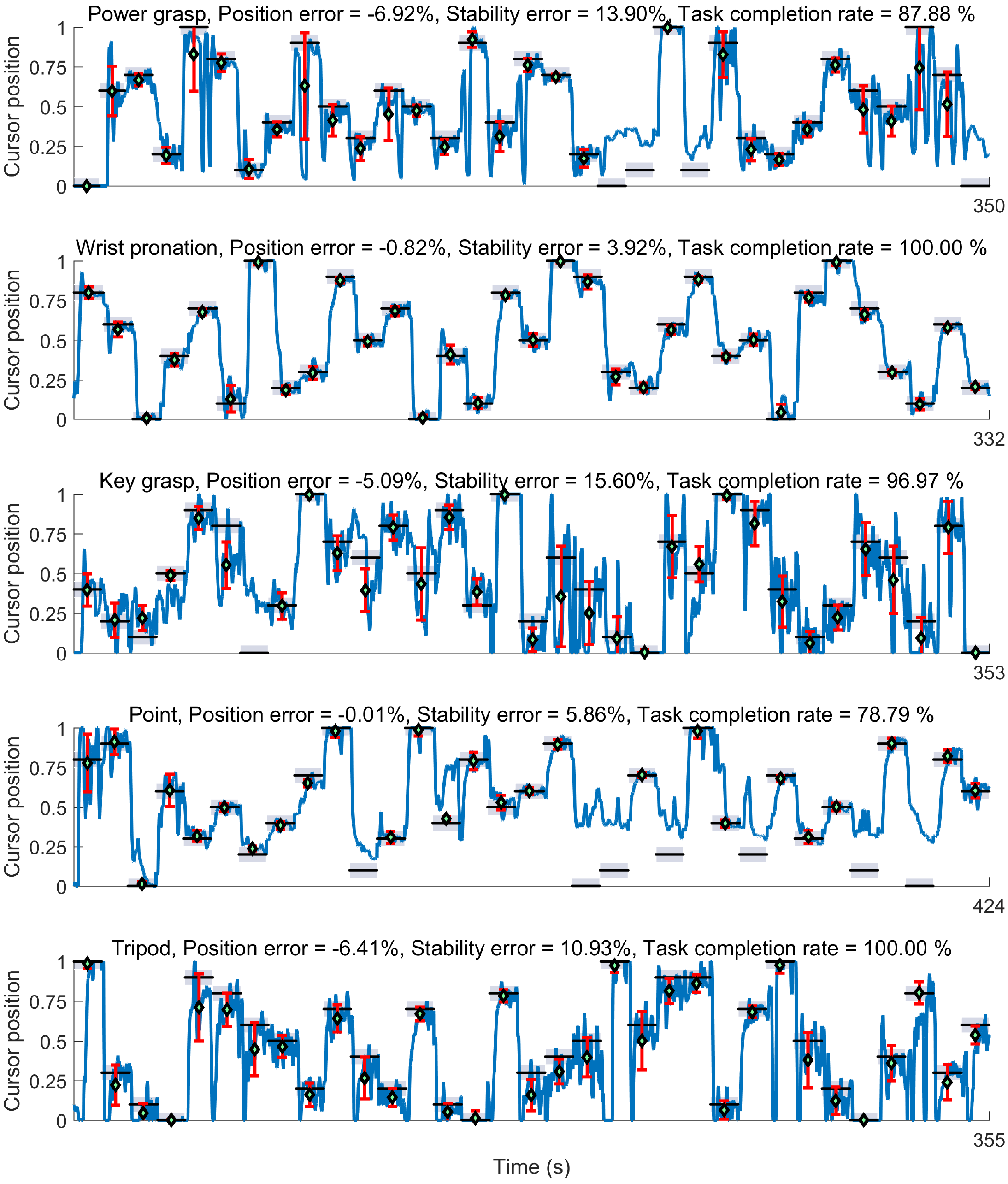}\\
\includegraphics[width=0.6\textwidth]{targetHoldingLegend}
\label{fig:am4TH}
\caption{Timeseries plots showing amputee subject \textit{Am5}'s performance in the target holding task. The user-controlled cursor position and target locations for five motions along with position and stability errors are shown. Motion legend- PG = \textit{power grasp}, WP = \textit{wrist pronation}, Po = \textit{point}, KG = \textit{key grasp}, Tr = \textit{tripod}.}
\label{fig:Am5TargetHolding}
\end{figure}

For the extended target holding task, \textit{point} was found to have the lowest average absolute position and stability error followed closely by \textit{wrist pronation} for amputee subjects. However, average task completion rate (Fig.~\ref{fig:compRateMotion}) for \textit{point} was 87.88\%, while that of \textit{wrist pronation} was 100\% across all trials. Similarly, \textit{power grasp} and \textit{tripod} had comparable position and stability errors, although the average task completion rate was 84.85\% for \textit{power grasp} and 100\% for \textit{tripod}. Fig.~\ref{fig:compRateTargetPos} also shows that there are small decreases in task completion rates at the lower quartile (0 to 0.25) and upper tenths (0.9 to 1.0) of the total motion completion range for amputees as well as able-bodied subjects. Timeseries plots of the cursor and target positions shown in Figs.~\ref{fig:Am4TargetHolding} and \ref{fig:Am5TargetHolding} also indicate that the unachieved targets were mostly presented in the latter half of the task. 

Finally, a Fitt's law analysis for the movement time required to achieve each target against its corresponding index of difficulty was performed and results are shown in Fig.~\ref{fig:aggregateStatsTH}. The index of difficulty ($ID$) for the task was defined as in equation.~\ref{eq:indexDiff}. 

\begin{equation}
\label{eq:indexDiff}
ID = log_{2}\left [\frac{D_{t}}{2|Q|}+1\right ]  
\end{equation}

Where, $D_{t}$ is the distance between subsequent targets and $Q$ is the target's quantization bound. Fig. \ref{fig:movementTime} shows that for both able-bodied and amputee subjects, mean movement time increases with increasing task difficulty across all motions as is expected in a human-computer interaction task. A regression analysis was performed of the mean movement times versus index of difficulty for both able-bodied ($R^{2} = 0.79$) and amputee subjects ($R^{2} = 0.72$). Throughputs were found to be slightly higher for able-bodied subjects at 1.35~bits/s  compared to 1.19~bits/s for amputee subjects. Movement time intercept (y-axis intercept) was also higher for able-bodied subject at 0.72~s compared to 0.44~s for amputees.


\section*{Discussion}

We have developed an ultrasound-based sensing approach for detection of volitional user movement intent and extraction of proportional control signals in response to muscle deformations in the forearm. Our proposed \textit{proprioceptive sonomyograpic control} strategy allows spatially resolved muscle activity detection in deep seated muscle compartments in the forearm, and extraction of proportional signals from individual functional muscle compartments enabling higher fidelity of control compared to traditional myoelectric sensing techniques. Our approach also enables true positional control of the end-effector device, as opposed to velocity control commonly implemented using myoelectric signals.

Our proposed approach can enable proportional control with multiple degrees of freedom as it can classify between different degrees of freedom as we have shown before\cite{akhlaghi2016real}. In this study, we demonstrated that simple classification algorithm, involving a 1-nearest neighbor classifier with a correlation-based distance metric was able to classify user-intended motions for both able-bodied and upper limb amputees. This image analysis pipeline makes our system agnostic to anatomical landmarks and removes the need for computationally expensive tracking algorithms~\cite{castellini2012using}. However, due to the nature of our ultrasound-based image analysis pipeline, we utilize a reference ultrasound image frame corresponding to \textit{rest} state to compute a correlation-based distance metric to other motion classes. We treat \textit{rest} as a separate motion class as opposed to the absence of a signal as in traditional myoelectric control paradigms~\cite{Powell2014}. We show that when \textit{rest} is excluded from classification there is a significant increase in classification accuracy across all subjects. We believe, this is due to natural postural variations in the relaxed state of the flexor muscles in the forearm. As a result, \textit{rest} image frames have a higher intra-class dissimilarity compared to other motion classes. However, when \textit{rest} is excluded, our approach achieves motion distinguishability that is comparable to current myoelectric pattern recognition (PR) systems performing similar motions~\cite{Powell2014,Resnik2018}. However, sonomyography was able to achieve this classification accuracy with less than an hour of training, compared to several weeks of training needed for PR to achieve comparable classification accuracy~\cite{Powell2014}. In future studies, we plan to utilize a single sEMG electrode in combination with sonomyography, and decode rest based on the absence of sEMG signal.

It is interesting to note that amputee subjects, \textit{Am1}, \textit{Am3} and \textit{Am4} are long-time myoelectric prosthesis users and reported having an advanced level of control over their residual musculature owing to daily use with their prosthesis. This prior experience and repeated practice may have had some influence leading to their initially high cross-validation accuracies as seen in Table \ref{table:ampCV}. Along the same lines, subject \textit{Am5}'s, slightly lower initial accuracy, in the first session (S1) (see Table \ref{table:ampCV}) may have been a result of limited exposure to a powered prosthesis. In the following session (S2), \textit{Am5} demonstrated a higher initial as well as average accuracy compared to his first session, indicating that the effect of training on our system may have been retained. This seems to suggest that sonomyography can leverage existing motor skills in current, experienced myoelectric prosthesis users without the need for extensive retraining. Additionally, \textit{Am5} later reported an improvement in his ability to control his existing myoelectric prosthesis due to a clearer understanding of his residual muscle deformation in the context of phantom digit movements. On the other hand, congenital subject \textit{Am2} was able to achieve high motion discriminability for four motions within approximately 30 minutes. This suggests that sonomyography can provide an intuitive control paradigm for traumatic amputees as well as congenital amputees who may either lack phantom limb sensations altogether or have limited context of muscular proprioception in their residuum. This is corroborated by contralateral arm demonstrations by unilateral amputees (see video in Supplementary material M2) showing that the movements perceived in their phantom limb closely resemble the intended motion. This one-to-one correspondence between residual muscle movement, resulting kinesthesia and perceived phantom sensation enables sensorimotor congruence in amputee subjects.  The effect of sensorimotor congruence extends to proportional control task performance as well (see video in supplementary material). We show that traumatic and congenital amputee subjects with no prior experience of using a sonomyography-based interface are able to demonstrate fine graded control of an end-effector controlled by muscle activity in the forearm. Position errors for amputee and able-bodied subjects were below 3.5\%, with an average task completion rate higher than 94\% for for 11 graded targets. We believe that this intuitiveness of our sonomyographic control paradigm is due to the direct mapping between the extent muscle deformation in the forearm and the derived position-based proportional signal. This is in contrast to traditional myoelectric control strategies that map muscle activation intensity to the velocity of a prosthetic device~\cite{hudgins1993new, Scheme2014}.

Earlier studies with sEMG electrodes integrated into prosthetic shells have shown that electrode shift, donning/doffing and arm position can have an adverse affect on long-term control reliability~\cite{Young2012,Hwang2017,jiang2013effect,Geng2012,Young2011}. In previous studies, we have evaluated the effect of user arm position on classification accuracy for able-bodied subject and demonstrated the robustness of our technique to such variations in a previous work~\cite{akhlaghi2016real}. Although, our image analysis pipeline does not rely on anatomical landmarks for classification or proportional control, major changes in transducer position will likely severely degrade both classification accuracy and proportional control performance and require retraining. Considering the short training regimes required for our approach, typically 10 minutes or less, a user-driven, structured training sequence may be a viable solution to retrain the classifier. However, minor variations in transducer location and orientation due to movement of the residuum inside a prosthetic shell can be mitigated in real-time by appropriate filtering and wavelet-based machine learning techniques \cite{Khan2017}.

This work establishes the feasibility of an ultrasound-based, noninvasive, muscle activity sensing modality for  potential use in real-time control of multiarticulated prosthetic arms for individuals with upper-extremity amputation. Clinical ultrasound devices continue to be miniaturized and handheld systems are commercially available. However, the current form factor of commercial systems are still to large to be readily be integrated with commercial prosthetic arms. We are currently developing custom miniaturized and low-power ultrasound imaging instrumentation \cite{Tarbox2017} that can be directly integrated into a prosthetic shell for continuous, \textit{in-vivo} monitoring of muscle activity and control of associated prosthetic devices. Our novel \textit{proprioceptive sonomyographic control} approach may provide a means to achieve intuitive proportional control in future prosthetic devices and potentially significantly lower the rate of device rejection.


\bibliography{bibliography}

 \section*{Acknowledgements}
 
This work is supported by multiple grants from: United States Department of Defense, Award Number: W81XWH-16-1-0722 and National Science Foundation, Grant Number: 1329829.

\section*{Additional information}

Some of the authors also participated as subjects in the study.

 \section*{Author contributions statement}

A.D., B.M., S.P., N.A., M.H-L., W.J. and S.S. conceived the experiment(s). A.D., B.M., S.P., and R.H. conducted the experiment(s), A.D., B.M., and S.P analyzed the results. A.D., B.M., S.P., G.L., and S.S wrote the manuscript. All authors reviewed the manuscript. 

\textbf{Competing interests}: The authors have no competing interests to declare.  


\end{document}